# Moving Object Based Collision-Free Video Synopsis

Anton Jeran Ratnarajah*, Sahani Goonetilleke†,Dumindu Tissera‡, Kapilan Balagopalan§ and Ranga Rodrigo¶
Department of Electronic and Telecommunication Engineering, University of Moratuwa, Sri Lanka Email:
*130514h@uom.lk, †130179h@uom.lk, ‡130600t@uom.lk, §130270e@uom.lk, ¶ranga@uom.lk

*Abstract*—Video synopsis, summarizing a video to generate a shorter video by exploiting the spatial and temporal redundancies, is important for surveillance and archiving. Existing trajectory-based video synopsis algorithms will not able to work in real time, because of the complexity due to the number of object tubes that need to be included in the complex energy minimization algorithm. We propose a real-time algorithm by using a method that incrementally stitches each frame of the synopsis by extracting object frames from the user specified number of tubes in the buffer in contrast to global energyminimization based systems. This also gives flexibility to the user to set the threshold of maximum number of objects in the synopsis video according his or her tracking ability and creates collision-free summarized videos which are visually pleasing. Experiments with six common test videos, indoors and outdoors with many moving objects, show that the proposed video synopsis algorithm produces better frame reduction rates than existing approaches.

## I. Introduction

There have been many approaches for summarizing a long video clip into a short clip through temporal reduction, to reduce the effort of browsing long videos. Fast forwarding [1] and optical flow-based motion analysis to select keyframes [2] are the two main early approaches. These approaches fail to summarize in a fast-moving video and can create visually uncomfortable video, when fast forwarding. The concept of video synopsis has overcome these problems, because it optimally reduces spatial redundancies when there are no temporal redundancies. Condensing a video by rearranging the foreground objects for fast browsing is video synopsis. This has attracted a lot of attention, since it optimally reduces spatial and temporal redundancy while the objects move at the same speed as in the original video. This approach was first presented by Rav-Acha *et al* [3]. In this approach of summarizing, the video synopsis itself a video expressing the dynamic movements of the scene, whereas the relative timing between the activities may change.

There are two approaches of creating a video synopsis, namely offline and online approach. Offline approach [3] has two phases of processing. In the first phase the video is scanned through in advance and both trajectories (tubes) and background are captured and stored. In the second phase, all the object tubes are rearranged together by minimizing a cost function. Since this method is complex in time and space, when processing a long video, online video synopsis approach is preferred. In this approach both phases get processed in parallel [4].

Common methods used to create online and offline video synopsis can be categorized based on whether the entire trajectory is used or not. Video synopsis by energy minimization [3] and video synopsis based on potential graph collision [5] are trajectory-based methods. Although trajectory-based methods can maintain chronological order and create highquality results, the existing approaches fail to create collision free online video synopsis in real time. Video synopsis based on maximum a posteriori probability estimation [6] is a nontrajectory based online video synopsis approach, that works in real time. This approach is burdened with the shortcoming of repeated appearance of the same object in a frame and ghost shadows which make the output video visually displeasing.

Our contributions are twofold. First, we present a trajectorybased video synopsis system that surpasses the existing nontrajectory-based Frame Reduction Rate (FRR), due to the ability to control the cluster size (no. of objects in the synopsis at a time) and incrementally stitching the frames in the synopsis, hence being flexible as well. Second, the proposed approach creates visually pleasing synopsis videos by detecting overlaps between objects being placed in a frame and thereby avoiding collisions between moving objects and maintaining the sequence of object movement closer to that of the original video. Due to the ability of our system to work with the user-specified cluster-sizes, there is flexibility for the user to control the number of objects in each frame in the synopsis that matches with his or her ability to analyze. The cluster size control also makes the algorithm less complex, thereby making it real time.

Since the online video synopsis approach removes the object tubes once they are stitched in the synopsis video, our approach manages memory efficiently in long videos. Processing time of each frame in our approach is positively correlated with the number of object tubes contained in a cluster, whereas the length of the synopsis video is negatively correlated with it.

As the quality of object tube generation is dependent on the accuracy of multiple object detection and tracking, the detection and tracking accuracy of the proposed approach has been tested with M-30 and M-30-HD videos in GRAM dataset [7]. Synopsis videos were generated with video datasets used by non-trajectory-based online video synopsis approach in [6] and the output results have been quantitatively and qualitatively evaluated.

The rest of the paper is organized as follows. Section II describes our approach of creating synopsis video. Experimental results on the proposed approach is described in Section III. Section IV discusses in detail, our approach based on the results. Section V concludes the paper.

## II. Proposed Methodology

### *A. Overview*

The aim of video synopsis is to summarize a long video into a short video clip by reducing spatial and temporal redundancies. The objects are represented as tubes in space-time volume. In our proposed approach the tubes are rearranged, such that the time occupied by all the tubes is minimized, whereas the space occupied at each time instant is maximized, without changing the actual spatial location of each tube, as shown in Fig. 1. This is achieved by a two-phase approach.

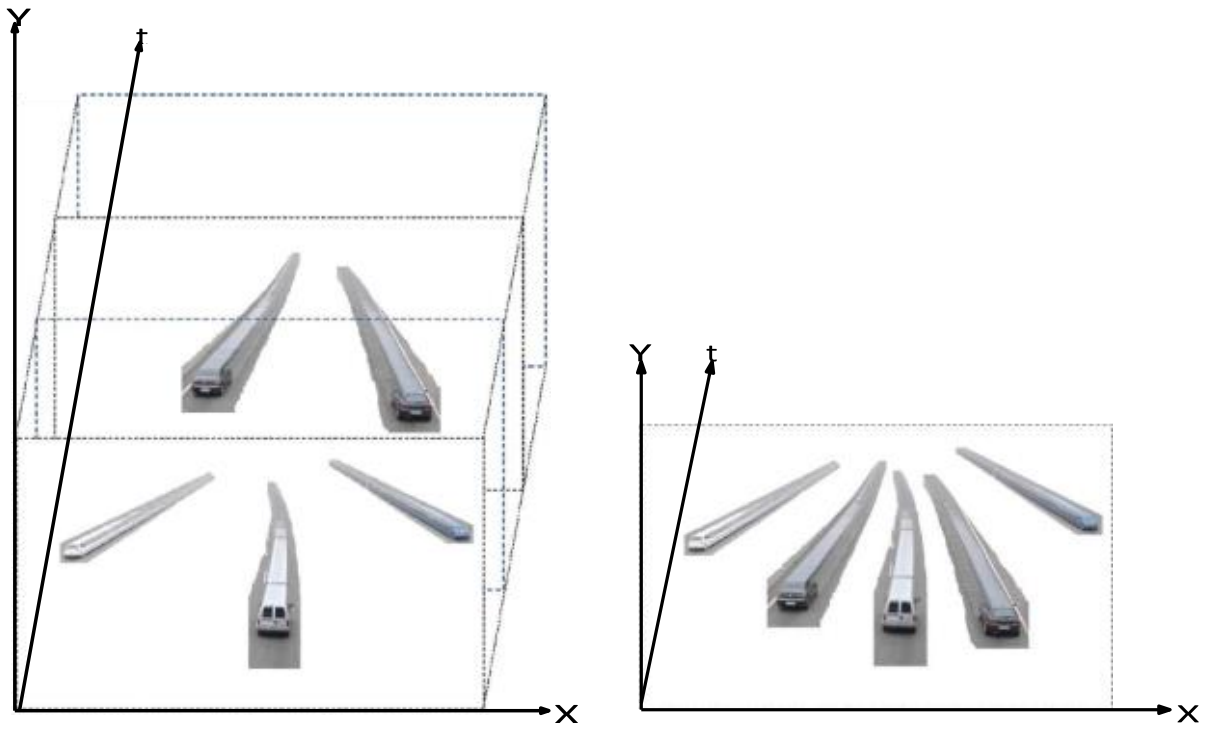

(a) Object tubes in original video. (b) Compact object tubes in synopsis video.

Fig. 1. Space-Time volume representation of object tubes in original M-30HD [7] video (left) and synopsis video (right).

In the first phase, foreground objects are extracted from the frame, localized and tracked to create object tubes as shown in Fig. 2. In the second phase, the created tubes are rearranged in parallel to create the synopsis video as shown in Fig. 3.

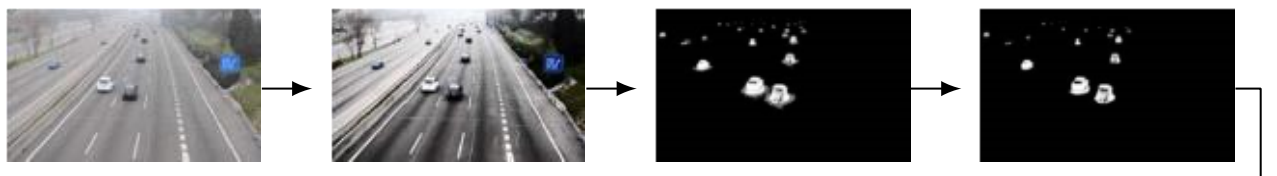

(a) Input. (b) Enhanced. (c) Background subtracted. (d) Shadow removed.

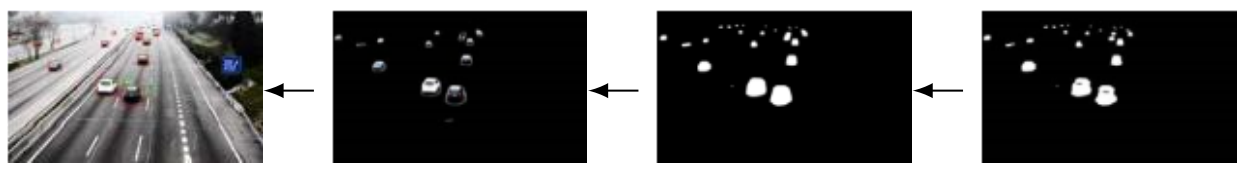

(e) Tracking same object in multiple frames. (f) Blob generation and extraction. (g) Convex hull generation. (h) After dilation and erosion.

Fig. 2. Proposed multiple object detection and tracking approach.

### *B. Moving Object Detection*

Since the primary purpose of moving object detection is to detect any non-stationary objects in a frame and object labels are not required, background subtraction method serves this purpose. Background subtraction can be carried out at pixel level or at region level. In the region-level approach [8], the Region of Interest (ROI) is divided into small blocks of interest and the background is modelled using the intensity variance of the blocks. Although this approach is low in computational complexity, it fails to separately detect multiple objects positioned nearby with a distinct bounding box.

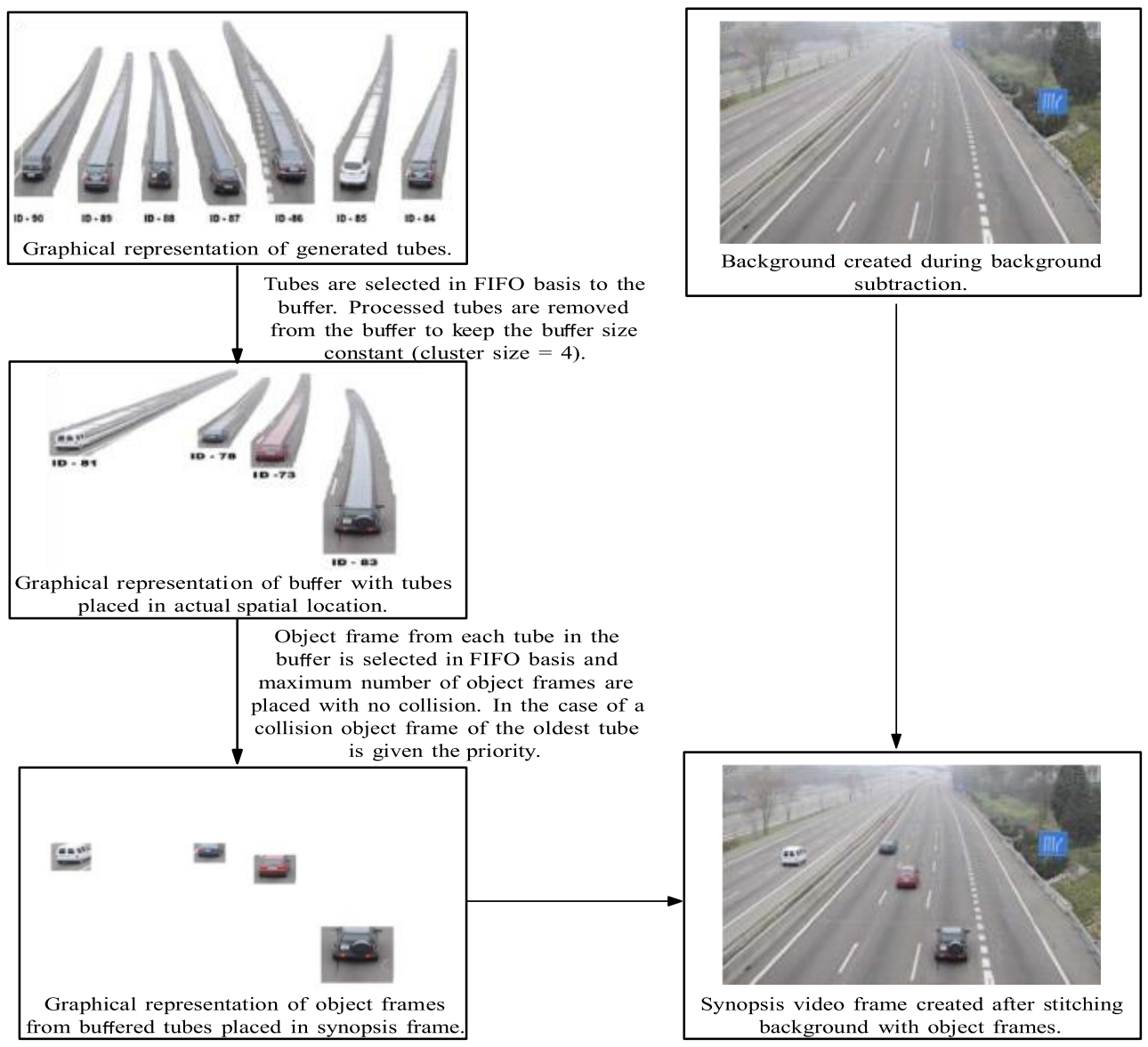

Fig. 3. Block diagram of multiple object tube re-arrangement and synopsis video generation in our approach.

A pixel-based approach is adapted to circumvent detection of numerous objects as a single object. This experiment uses Mixture of Gaussian (MOG) based approach implemented in OpenCV. Pixel-based techniques mentioned in [9] and [10] are used to segment background and foreground based on MOG. Here the number of mixture components has been determined dynamically per pixel. In the experiment, the last 100 samples are used as training data and the density is reestimated whenever a new sample replaces the oldest sample in the training data. In the above implementation, a shadow is detected if the pixel is darker than the background. Here the shadow threshold of 0.5 is set, implying that if a pixel is more than twice darker, it is not considered as a shadow. Optimum variance threshold for the pixel-model match is set to 25 for the experimented video datasets.

Once the foreground and background have been segmented, a binary image is created by assigning one for foreground pixels and zero for background pixels. Then, the binary image is dilated and eroded to close holes in the foreground image and to remove noise. Next, contours are generated by joining all continuous points along the boundary of each foreground object. Using these contours, convex hulls are generated for each foreground object and then each moving object is segmented and represented as the smallest rectangular box enclosing the convex hull to create tubes (See Fig. 2).

### *C. Multiple Object Tracking*

Multiple object tracking comprises of 2 key stages, namely object detection and motion prediction. According to [11] multiple object tracking can be classified as recursive and non-recursive. This paper proposes a recursive method where the current state information is estimated using only previous frame information. The existing works [7], [12] use Extended Kalman Filter approach to predict the motion of objects. This paper presents a less complex motion prediction approach that performs well in highway vehicle tracking and at satisfactory accuracy when objects move with non-linear motion. In this approach, for each tracked object, the center of the bounding box in last 10 frames is used to predict the new position of that object. Optimal weights and number of object frames used for prediction are selected by experimenting on GRAM dataset [7] and evaluating the accuracy of tracking algorithm.

Let $i$ be the current object frame number, $C[i]$ be the center of the tracked object in $i^{th}$ frame, $P[i]$ be the predicted center of the tracked object and $D[i]$ be the difference between predicted center and current center.

If

$$i \leq 10, \quad D[i] = \frac{\sum_{n=1}^{i-1} (C[n+1] - C[n]) \times (n)}{\sum_{n=1}^{i-1} (n)} \tag{1}$$

else

$$D[i] = \frac{\sum_{n=1}^{9} (C[i+1-n] - C[i-n]) \times (10-n)}{\sum_{n=1}^{9} (n)} \tag{2}$$

$$P[i] = C[i] + D[i] \tag{3}$$

After predicting the future center positions of currently tracked objects, this approach maps them with the future detected objects. As one of the characteristics of tracking is to remove noisy detection, this approach gives an object identity number to detected tubes only if the objects have been consecutively tracked through at least one second.

### *D. Object Based Video Synopsis*

In the online approach of object based video synopsis, both tube generation and tube re-arrangement occur in parallel. While multiple objects are being tracked in each frame, the object tubes are generated and the background image is stored. In this approach, tubes are re-arranged while new tubes are generated in parallel. Here the user defines the size of the cluster of tubes to be processed at a given time. Synopsis video is created by placing the maximum possible number of objects in each synopsis frame in their chronological order, subject to zero collision of tubes. In this approach the tubes are placed in the synopsis video in first-in, first-out, basis. The algorithm used for tube re-arrangement and synopsis video generation is given in Algorithm 1.

Let GTB be the generated tube buffer, CTB be the cluster tube buffer, OF be the object frame and CS be the cluster size.

Algorithm 1 Tube re-arrangement and synopsis video generation.

```
while 1 do
  n ← n + 1
  while CTB.size() < CS do
  CTB.add(GTB[1])
      GTB[1].remove() end
  while
  for i ← 1,CS do
      if (CTB[i].OF[1] ∩ (CTB[i − 1].OF[1] ∪ · · · ·
∪CTB   .OF[1] ∪ CTB[1].OF[1])) = 0 then
[2]
  SynopsisFrame[n] ← CTB[i].OF[1]
      CTB[i].OF[1].remove() end if
  end for
  for i ← 1,CS do
  if CTB[i].size() = 0 then
      CTB[i].remove() end if
  end for
  end while
```

## III. RESULTS

To evaluate the performance of the proposed video synopsis algorithm, we used the video datasets in [6] and [7] . Since the accuracy of tube generation depends on how well the proposed methodology detects and tracks multiple objects, we evaluated the tracking accuracy within the ROI of the annotation of GRAM dataset [7]. Table I shows detailed information of GRAM dataset.

TABLE I
DETAILED INFORMATION ABOUT GRAM DATASET.

| Video Name | M-30-HD | M-30 |
|---|---|---|
| Size of the image (96 dots per inch) | 1200 ×720 | 800×480 |
| Total Number of Vehicle | 241 | 270 |
| Frames per second | 30 | 30 |
| Total Number of Frames | 9310 | 7520 |
| Weather Condition | Cloudy | Sunny |

We used the code provided by GRAM dataset [7] to evaluate the accuracy of the tracking algorithm. It calculates the average precision to determine the accuracy of the tracking algorithm and plots the precision vs. recall curve. The definition of precision and recall used here are as follows:

Let $TP[n]$ be the true positive in $n^{th}$ frame, $FP[n]$ the false positive calculated using false detection and multiple detection of same object in $n^{th}$ frame, $NP[n]$ the total number of annotated detections in $n^{th}$ frame, $N$ the total number of frames in the video and $i$ the $i^{th}$ frame.

$$\text{Precision} = \frac{\sum_{n=1}^{i} TP[n]}{\sum_{n=1}^{i} (TP[n] + FP[n])} \tag{4}$$

TABLE II
AVERAGE PRECISION OF PROPOSED AND EXISTING METHODS.

| Detection Method | Tracking Method | M-30-HD | M-30 |
|---|---|---|---|
| Multi Time Spatial Image Based Vehicle Detection [13] | KF | 0.478 | 0.291 |
| | PF | 0.681 | 0.664 |
| | MIKF | 0.806 | 0.741 |
| | MIPF | 0.769 | 0.701 |
| HOG [7] | EKF | 0.524 | 0.3009 |
| Proposed Method | Proposed Method | 0.871 | 0.799 |

$$\text{Recall} = \frac{\sum_{n=1}^{i} TP[n]}{\sum_{n=1}^{N} NP[n]} \tag{5}$$

$$\text{Average precision} = \text{Precision} \times \text{Recall} \tag{6}$$

Table II provides a comparison of the average precision values calculated using different multiple object detection and tracking approaches with the proposed approach. Since the higher average precision value relates to higher accuracy, it can be seen that the proposed method detects and tracks accurately.

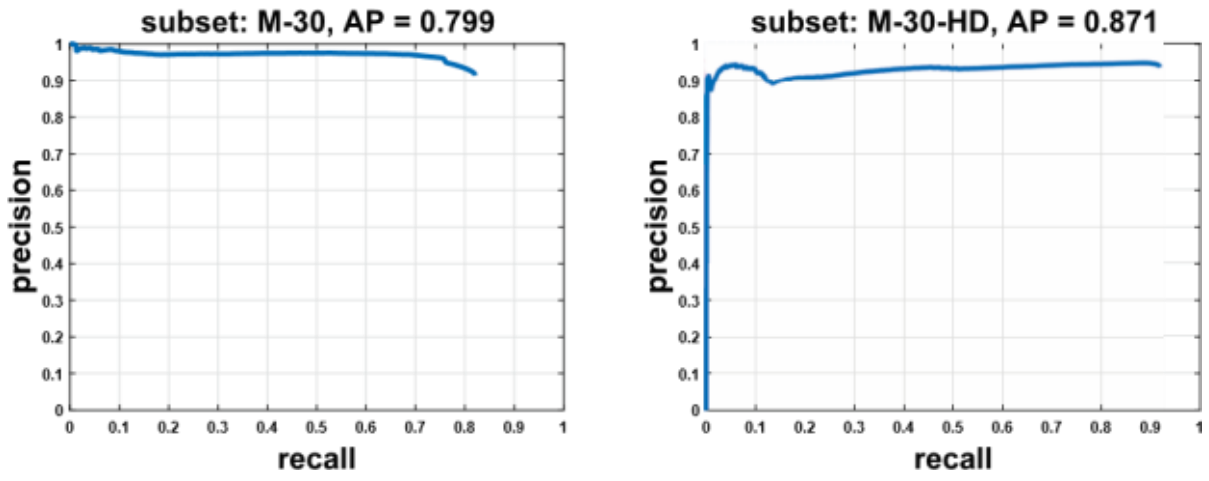

(a) M-30. (b) M-30-HD.

Fig. 4. Precision-recall curve for tracking M-30 and M-30-HD videos.

Fig. 4. illustrates the precision vs. recall curve for both the videos. The graphs indicate that the precision of the algorithm is more than 90%. Also, it can accurately detect more than 80% of the annotations in M-30 video and more than 90% of that in M-30-HD video. Since we use first 50 frames to initially train the background model and rerun the video for evaluation purpose, we see poor results in the beginning of the graph (M-30-HD video). Once the background model becomes stable over time, consistent results can be noted.

The significance of the proposed approach is that it can detect and track a larger ROI than the annotation of the above videos. Fig. 5 depicts that the proposed approach detects and tracks accurately within the ROI of the annotation, while it also can detect and track outside the ROI.

The video synopsis algorithm was tested on the four videos used in [6]. The detailed information of the dataset is shown in Table III. We ran experiments on different cluster sizes (CS) of the tubes and the results are in Table IV. Here, FRR is calculated by dividing the Total number of frames in the Synopsis Video (TSV) by the Total number of frames in the Original Video (TOV). Frames per Second (FPS) is calculated by dividing the TOV by the total time taken to create the synopsis video.

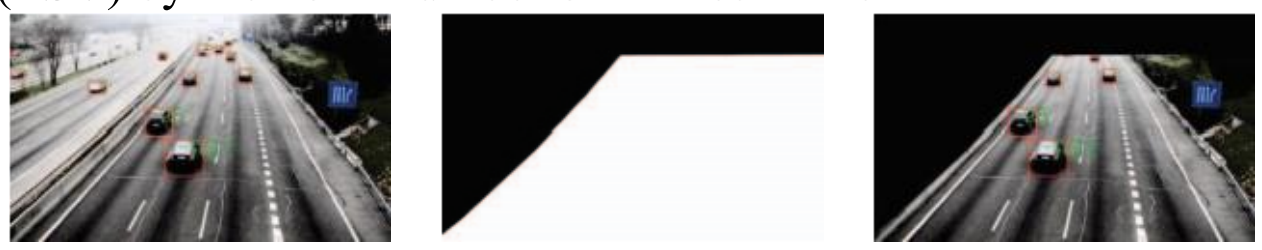

(a) (b) (c)

Fig. 5. Multiple detection and tracking of 1000th frame in M-30-HD video (a), ROI of annotation (b), multiple detection and tracking within the ROI (c).

TABLE III
DETAILED INFORMATION ABOUT SYNOPSIS DATASET.

| Video | Duration | # of frames | FPS | # of objects |
|---|---|---|---|---|
| Cross Road (V1) | 01:04:59 | 70195 | 18 | 1677 |
| Street (V2) | 01:13:33 | 79449 | 18 | 871 |
| Hall (V3) | 01:01:49 | 66771 | 18 | 276 |
| Sidewalk (V4) | 00:58:18 | 104864 | 30 | 334 |

TABLE IV
FPS AND FR FOR DIFFERENT CLUSTER SIZE FOR DATASET VIDEOS.

| CS | V1 | | V2 | | V3 | | V4 | |
|---|---|---|---|---|---|---|---|---|
| | FR | FPS | FR | FPS | FR | FPS | FR | FPS |
| 10 | 0.211 | 56.9 | 0.213 | 43.2 | 0.121 | 85.9 | 0.120 | 89.3 |
| 20 | 0.167 | 58.0 | 0.141 | 43.0 | 0.088 | 85.0 | 0.095 | 87.3 |
| 40 | 0.142 | 56.6 | 0.112 | 42.9 | 0.073 | 83.7 | 0.084 | 84.3 |
| 200 | 0.121 | 55.1 | 0.095 | 40.5 | 0.070 | 67.4 | 0.078 | 72.0 |
| 1000 | 0.119 | 46.2 | 0.095 | 35.1 | 0.070 | 57.5 | 0.077 | 56.2 |

Through the experiment, it can be concluded that time taken to produce the synopsis video is directly proportional to synopsis video size, cluster size and average object density in the original frames. Although FRR is inversely proportional to the cluster size, the synopsis video becomes unpleasant to watch for large cluster sizes, since there would be more flickering in the video to avoid occlusions and all the objects would be tightly packed in the video.

TABLE V
COMPARISON OF TOTAL NUMBER OF FRAMES IN SYNOPSIS VIDEO (TSV) AND FRAME REDUCTION RATE(FR).

| | TOV | [14] | | [6] | | Proposed Method | | |
|---|---|---|---|---|---|---|---|---|
| | | TSV | FR | TSV | FR | TSV | FR | CS |
| V1 | 70195 | 12685 | 0.181 | 7876 | 0.112 | 12906 | 0.184 | 15 |
| V2 | 79449 | 18703 | 0.237 | 21371 | 0.269 | 16947 | 0.213 | 10 |
| V3 | 66771 | 14379 | 0.215 | 11271 | 0.169 | 7311 | 0.11 | 12 |
| V4 | 104864 | 18250 | 0.174 | 17340 | 0.165 | 15399 | 0.147 | 7 |

After the experiment, the optimal cluster size for the four videos were chosen based on the creation of visually pleasing synopsis videos, FPS and FRR. The proposed method has been compared with existing methods which used the above dataset and the results are tabulated in Table V. We can observe that the proposed method has lower FRR for V2, V3 and V4 videos at the optimal cluster size. This implicates that the proposed work can compress better, while preserving all important information in the synopsis video. As the vehicles move in different directions within the same region in V1, our approach has slightly higher FRR to avoid collision.

TABLE VI
FRAMES PER SECOND (FPS) AND SIZE UNDER H.264 COMPRESSION (BYTES).

| Video | FPS | | Cluster Size | Size in Bytes | |
|---|---|---|---|---|---|
| | Original | Synopsis | | Original | Synopsis |
| V1 | 18 | 57 | 15 | 49.9M | 25.7M |
| V2 | 18 | 43.23 | 10 | 41.8M | 21.5M |
| V3 | 18 | 85.55 | 12 | 26.9M | 11.3M |
| V4 | 30 | 89.82 | 7 | 76.6M | 16.8M |

We used an Intel core i7-4770 CPU @ 3.4GHz for the experiment. Table VI shows that the proposed method runs in real time for less dense video datasets. Another use of the creation of synopsis video is that the original video can be compressed. Since CCTV cameras record almost continuously, a large amount of space is needed for storage. Further, as video synopsis compresses the video with only useful information, space can be efficiently managed. Table VI compares the size of original and synopsis video under H.264 compression. From that we can decipher that a video can be efficiently compressed to less than 50% of the original size on average.

Fig. 6 illustrates the synopsis video created using four videos in synopsis dataset. Each moving object in the synopsis video is labeled with the time it appears in the original video. As the group of objects moving together are localized and tracked together, the produced synopsis video preserves important information. The output videos are available at

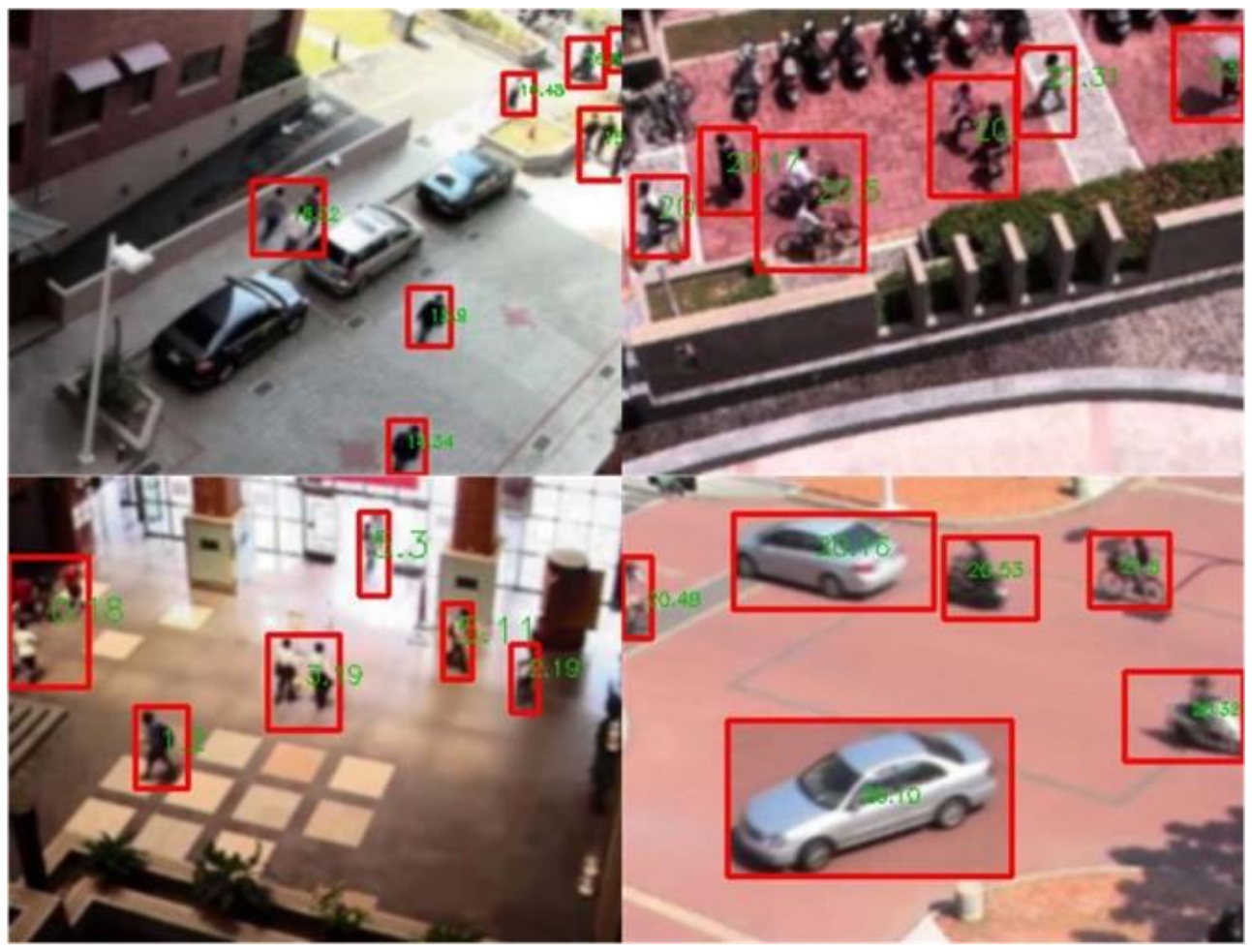

Fig. 6. Synopsis video created using the dataset.

https://anton-jeran.github.io/M2SYN/

## IV. DISCUSSION

### A. Cluster Size

Cluster size determines the number of tubes used to create the synopsis video at each instance. It also thresholds the maximum number of objects in a synopsis video frame. In Fig. 7, for small cluster sizes, such as 5, majority of the space is vacant in the synopsis video. This results in the creation of a synopsis video of large duration. When the cluster size is very large, such as 1000, ROI occupied by moving objects is completely packed and there is a higher probability that those object frames of the tubes would not be placed continuously due to different trajectories followed by different tubes. This results in flickering in the synopsis video. A cluster size of 20 produces optimal synopsis video for ”Cross Road” video with less duration and a reasonable level of flickering. As the optimal cluster size is totally subjective and it depends on the average object size relative to the size of the frame and motion velocity, it should be set as a variable parameter.

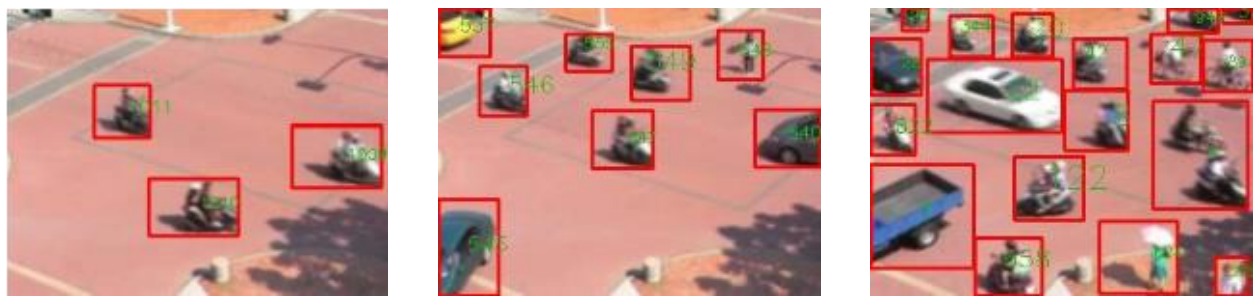

(a) Cluster Size = 5 (b) Cluster Size = 20 (c) Cluster Size = 1000

Fig. 7. Synopsis video created for Cross Road video at different cluster size.

### B. Failure Cases

Due to the placement of objects from different time instances in a single frame in the synopsis video and the asynchronous update of the background over time, the following fault cases may arise.

1) *Objects Overlap:* Since objects once stationary may blend into the background with time, moving objects may be placed over the objects that have become background. This may give faulty impressions to the user that two objects have overlapped. Fig. 8(e) depicts an instance where the black car has become background over time when it was parked. The person who has walked through the area covered by the black car when it is not present in the original video is depicted to walk over the car in the synopsis video.

2) *Ghost Movement:* This arises due to the number of frames used to train the Gaussian model for detection is limited to, eg. 100 frames in an 18 FPS video. Any unusually slow movements such as parking a car will be suddenly depicted in the synopsis video through the background update. Fig. 8(a) to Fig. 8(d) show a car parking scenario covered by background update.

*3) Multiple Instances of the Object:* As an object may arrive through the background update before its corresponding object frame is stitched, the same object's multiple instances may be observed in a single frame. Fig. 8(f) shows corresponding instance of the black car.

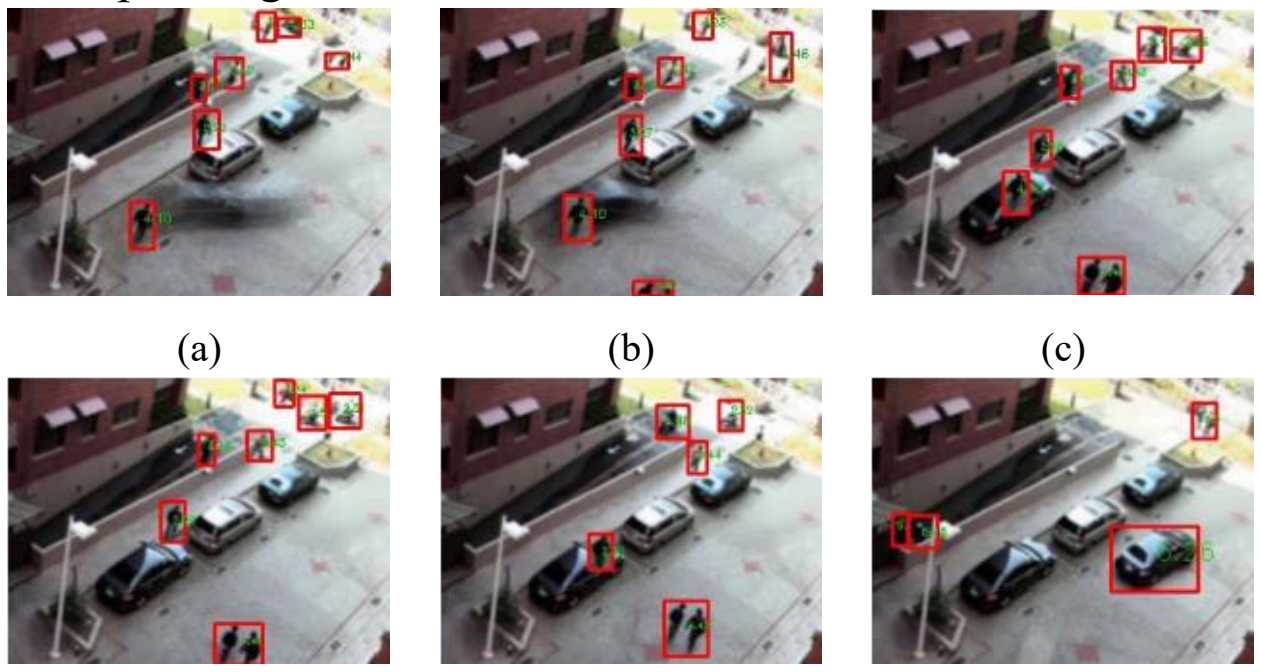

(a) (b) (c)

(d) (e) (f)

Fig. 8. Fault case in Street Video.

### C. Real Time

Despite of faulty cases in situations which has less probability of occurrence, the proposed algorithm runs at higher FPS than the recorded rate. Therefore, the proposed algorithm can run in real time even for higher density videos. This is very useful, since CCTV cameras record videos almost continuously, video synopsis should run in real time, to synchronously summarize in real usage without any lag accumulation.

### D. Highways

The proposed approach works well in highways at large cluster sizes, since all the vehicles in a lane follow same trajectories and there is no issue of the vehicle becoming background in normal situations, as they are fast moving. Fig. 9 shows synopsis video created at cluster size of 50 and 100 in M-30 and M-30-HD videos respectively.

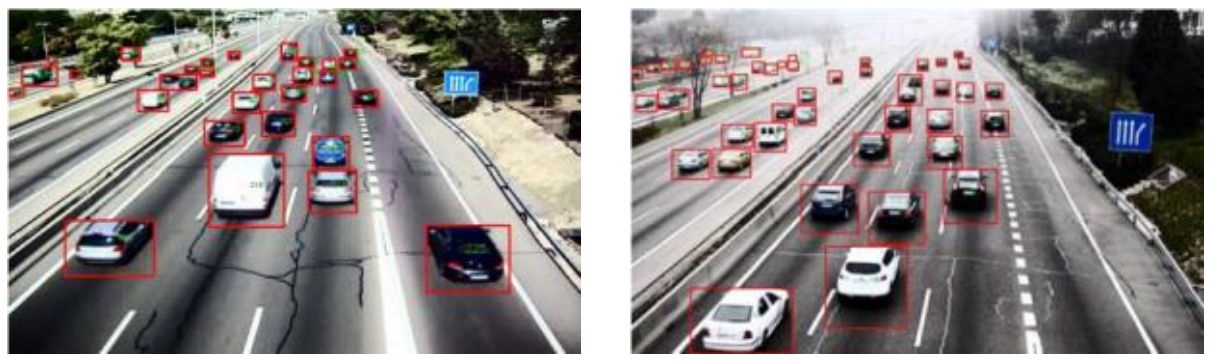

(a) M-30 synopsis video at cluster (b) M-30-HD synopsis video at size of 50 cluster size of 100

Fig. 9. Synopsis video created with GRAM dataset.

## V. CONCLUSION

This paper proposes a less complex, collision free, trajectory-based online video synopsis algorithm, which can process as fast as non-trajectory-based online video synopsis algorithms. The results show that our approach has a better frame reduction rate than existing approaches and can be processed in real time. The proposed approach focuses on producing a visually-pleasing summarized video for the user. Therefore, we adopted strategies such as making a collisionfree video. The proposed approach gives flexibility to the user to limit the maximum number of objects in a synopsis frame. This allows the user to set values based on his or her tracking ability in the summarized video. This paper has also qualitatively and quantitatively discussed the effects of very low and very large thresholding values.

The proposed algorithm has been tested on six videos in the GRAM and Synopsis datasets as well as on other videos. We verified the accuracy of the tracking algorithm and efficiency of the video synopsis algorithm using the test videos. As the dataset videos are taken indoors as well as outdoors, covering different scenarios, the proposed algorithm is verified to work under different conditions.

In this paper we discussed the problems that occur due to multiple objects from different time frames being stitched on the background which is updated asynchronously. This problem is expected to be overcome in future work.